\pdfoutput=1
\documentclass[a4paper,11pt,onecolumn]{personalarxiv}
\usepackage{mathtools,needspace}
\usepackage{tikz}
\usetikzlibrary{arrows.meta,positioning,fit,calc}
\usepackage{placeins}
\newcommand{\V}{\mathcal V}
\newcommand{\Qone}{\mathcal Q_1}
\newcommand{\E}{\mathbb E}
\newcommand{\KL}{D_{\mathrm{KL}}}
\newcommand{\ds}{\textsc{DSpark-style}}

\DeclareMathOperator{\softmax}{softmax}
\newtheorem{proposition}{Proposition}

\theoremstyle{definition}
\newtheorem{assumption}{Assumption}
\newcolumntype{Y}{>{\raggedright\arraybackslash}X}
\newenvironment{sampletext}{\leavevmode\par\vspace{2pt}\begingroup\linespread{1}\small\selectfont\raggedright\setlength{\parindent}{0pt}\setlength{\parskip}{4pt}\leftskip=8pt\rightskip=8pt}{\par\endgroup}
\definecolor{draftgray}{RGB}{244,245,247}
\definecolor{draftink}{RGB}{57,63,73}
\definecolor{localblue}{RGB}{50,85,128}
\definecolor{stateblue}{RGB}{230,238,248}
\definecolor{globalgreen}{RGB}{38,103,91}
\definecolor{stategreen}{RGB}{230,242,237}

\title{The Generative Gap Between Speculative Drafter and Continuous Diffusion Language Model}

\author[1,2]{Yujian Chen}
\author[1,3]{Hao Jiang}
\author[5]{Guang Cheng}
\author[4]{Weijian Luo\thanks{Corresponding author and Project Leader: \href{mailto:pkulwj1994@icloud.com}{\texttt{pkulwj1994@icloud.com}}}}
\author[1,3]{Tianyang Hu\thanks{Corresponding author: \href{mailto:hutianyang@cuhk.edu.cn}{\texttt{hutianyang@cuhk.edu.cn}}}}

\affil[1]{Polyadic Research Group}
\affil[2]{School of Computer Science, Peking University}
\affil[3]{SAGA Lab, The Chinese University of Hong Kong, Shenzhen}
\affil[4]{Dots Studio, Xiaohongshu}
\affil[5]{The University of California, Los Angeles}
\paperpdfauthor{Yujian Chen, Hao Jiang, Weijian Luo, Guang Cheng, Tianyang Hu}

\begin{document}
\maketitle

\begin{abstract}
Continuous diffusion language models (CDLMs) and speculative drafting offer two approaches to efficient language generation, but their generative capabilities are typically evaluated under different conditions.  
CDLMs generate independently through iterative refinement, whereas speculative drafters produce proposals that a target model verifies and corrects. 
How do their underlying architectures compare when both must model a text distribution without verification? 
We instantiate CDLMs with LangFlow and speculative drafting with DSpark, adapting the latter’s Markov and recurrent-head architectures for standalone generation. 
We train all models from scratch for unconditional, single-block OpenWebText generation at lengths 16, 128, and 1024, without target features or verification, using common data preprocessing, backbone dimensions, and training-update budgets.
We evaluate generation perplexity (GenPPL) under an external language model and distributional similarity to reference text using MAUVE. LangFlow outperforms both DSpark variants on both metrics at lengths 128 and 1024 with its default denoising budget equal to sequence length, while DSpark-RNN leads at length 16.
To understand this comparison, a conditional KL analysis characterizes dependencies excluded by the Markov head’s previous-token conditioning. 
Recurrent memory removes this restriction and improves both metrics, but leaves the observed longer-sequence gap. 
Teacher probability matching further lowers Markov GenPPL without an observed MAUVE gain. 
In contrast, increasing LangFlow’s denoising steps at fixed weights improves both metrics at lengths 16 and 128, changing the length-128 ordering relative to DSpark-RNN. 
Together, these results establish a longer-sequence advantage for LangFlow under the evaluated budgets and highlight iterative full-sequence refinement as a route to improving standalone generation beyond the gains obtained from recurrent head memory.


\end{abstract}

\section{Introduction}
\label{sec:introduction}

Parallel language generation seeks to reduce sequential computation while preserving dependencies among tokens.
Continuous diffusion language models (CDLMs) \citep{li2022diffusionlm,strudel2022sed} and speculative drafting offer two approaches to this goal.
CDLMs generate text through iterative refinement, whereas speculative drafters propose tokens that a target model verifies and corrects \citep{leviathan2022speculative,chen2023sampling}.
Their usual evaluations therefore place different demands on the generator: a standalone model must itself produce coherent samples, while a drafter can rely on target correction.
This difference makes it difficult to assess their relative generative capabilities from performance in their original settings.
We ask: \emph{how do their underlying architectures compare as standalone generators, and what benefits does iterative full-sequence refinement provide?}

We instantiate CDLMs with LangFlow \citep{chen2026langflow} and speculative drafting with DSpark \citep{cheng2026dspark}, adapting the latter's Markov and recurrent-head architectures for standalone generation.
For a standalone generator, an early choice of entity or topic should inform what follows.
We focus on how information about such choices enters the generation process.
LangFlow repeatedly refines a representation of the entire sequence, allowing positions to exchange information before tokens are finalized.
The DSpark variants compute backbone features once and incorporate generated tokens through a lightweight output head.
The Markov head accesses only the preceding token, while the recurrent head can retain information from earlier history.
These mechanisms provide different ways to coordinate choices across a sequence.

We compare the models on unconditional, single-block OpenWebText generation \citep{Gokaslan2019OpenWeb}.
Without a semantic prompt or previously generated blocks, each model must establish both its content choices and the dependencies that follow from them.
We train separate models from scratch at each sequence length, without target features or verification, using common data, backbone dimensions, and training-update budgets.
The drafting-style architectures therefore learn directly for the standalone task.
We retain their different generation procedures: one backbone pass for the drafters and iterative denoising for LangFlow.
Generation quality is assessed using generation perplexity (GenPPL) under an external language model and distributional similarity to reference text using MAUVE \citep{pillutla2021mauve}.

Under the evaluated sampling budgets, LangFlow outperforms both DSpark variants on both metrics at lengths 128 and 1024, while DSpark-RNN leads at length 16.
Recurrent memory improves the drafter at every tested length, but does not close the longer-sequence gap.
This pattern motivates two questions: what limitations of the drafter's access to history can be identified analytically, and how can generation be improved beyond extending that access?

Our analysis identifies a specific limitation of the Markov head: information about an earlier choice may no longer be recoverable from the preceding token.
We characterize the resulting restriction on the distributions it can represent and show how recurrent memory removes this particular obstacle.
This separates a representational limitation of the Markov interface from the empirical question of how effectively a recurrent model learns to use history.
We then examine whether richer supervision helps within the same Markov architecture.
Matching a teacher's next-token probabilities changes the training targets without providing additional access to generated history.
It lowers GenPPL without an observed MAUVE gain, showing that better evaluator predictability does not necessarily translate into better distributional similarity.

LangFlow offers another improvement route through additional refinement during generation.
We examine this route by increasing denoising steps while keeping the trained model weights fixed.
Both metrics improve in the sampling-budget study, and additional denoising allows LangFlow to overtake DSpark-RNN at length 128.
The recurrent drafter retains its short-sequence advantage throughout the tested budgets.
These results show that the comparison depends on how much computation is used during sampling, as well as on the learned model.
They highlight a benefit of iterative continuous generation: the same trained model can improve its outputs through additional full-sequence refinement.

Our contributions are:
\begin{itemize}
    \item A standalone comparison that characterizes the generative gap between CDLM and drafting-inspired architectures across sequence lengths (Section~\ref{sec:main-results}).
    \item An analysis of how access to generated history shapes the dependencies that a sampling head can represent (Sections~\ref{sec:formulation}--\ref{sec:structure}).
    \item Evidence distinguishing the effects of teacher supervision and generation-time computation, including fixed-weight improvements from LangFlow's iterative refinement (Sections~\ref{sec:teacher-results}--\ref{sec:block-results}).
\end{itemize}

\section{Background and Related Work}
\label{sec:related}

\paragraph{Drafting and verification.}
Speculative sampling preserves the target distribution through correction \citep{leviathan2022speculative,chen2023sampling}. Medusa uses parallel heads \citep{cai2024medusa}, EAGLE-3 predicts tokens from fused target features \citep{li2025eagle3}, and DFlash uses block diffusion \citep{chen2026dflash}. Target-assisted results do not measure standalone generation.

\paragraph{Dependencies within proposals.}
DSpark adds Markov or recurrent heads \citep{cheng2026dspark}; Domino uses recurrent causal correction \citep{huang2026domino}. xPress addresses the marginal--conditional mismatch through parallel Jacobi causal refinement \citep{wang2026xpress}. DDTree and PRESTO explore candidate trees, with PRESTO aligning ranking with prefix-based verification \citep{ringel2026ddtree,wang2026presto}. We examine these dependencies in standalone generation.

\paragraph{Parallel readout and shared state.}
Latent fertility \citep{gu2017nonautoregressive}, iterative refinement \citep{lee2018refinement}, and Mask-Predict \citep{ghazvininejad2019maskpredict} coordinate alternative outputs. Shared state can couple tokens despite parallel readout.

\paragraph{Continuous representations and token recovery.}
Diffusion-LM and Self-conditioned Embedding Diffusion develop embedding-space denoising \citep{li2022diffusionlm,strudel2022sed}; Plaid develops likelihood-based modeling \citep{gulrajani2023plaid}. LangFlow and RePlaid revisit schedules, representations, and architectures \citep{chen2026langflow,yang2026replaid}. CoDAR, LDLM, and Cola DLM respectively study contextual token recovery, joint latent learning, and semantic priors \citep{shen2026codar,meshchaninov2026ldlm,guo2026cola}; decoder-interface analysis emphasizes recoverability beyond reconstruction error \citep{du2026interface}.

\paragraph{State space and iteration budget.}
Discrete diffusion extends from SEDD/MDLM \citep{lou2024sedd,sahoo2024mdlm} to LLaDA/Dream \citep{nie2025llada,ye2025dream}. Block Diffusion combines between-block autoregression with within-block diffusion \citep{arriola2025block}. Conversely, continuous FLM's distilled FMLM supports one-step generation \citep{lee2026flow}, and FlowLM adapts diffusion for few-step flows \citep{zhang2026flowlm}. Representation, factorization, and iteration are distinct choices; our NFE findings concern the evaluated LangFlow.

\paragraph{Evaluating joint generation.}
GenPPL can reward poor coverage even with an entropy criterion \citep{franca2026hacking}. MAUVE estimates distributional divergence \citep{pillutla2021mauve}. We combine these measurements with supplementary lexical diagnostics, distinguishing standalone quality from target-verified acceptance.

\section{Standalone Generation and Sampling Interfaces}
\label{sec:formulation}

We first define the standalone generation task, then specify what sampled history each drafter head can use. These interfaces set up the information-access result in Section~\ref{sec:structure} and the architectural comparison in Section~\ref{sec:continuous}.

\subsection{Comparison setting}
Let $X_{1:B}\in\V^B$ be a block of $B$ tokens and let $C$ contain all information available before that block is sampled. A generator defines $q_\theta(X_{1:B}\mid C)$; $p(X_{1:B}\mid C)$ denotes the reference distribution. In the unconditional single-block setting, $C=c_0$ contains fixed control inputs, such as a BOS token and the chosen length, but no sample-specific semantic prefix. Independent runs must generate both a content choice and its subsequent consequences.

We distinguish the generation block size $B$, the total output length $L$, and the training length distribution. A long document may be produced by concatenating small blocks, as in Block Diffusion \citep{arriola2025block}, so support for large $L$ does not imply that the same model represents a rich distribution within one large block. When generation proceeds over multiple blocks, prior generated text becomes part of $C$. All within-block statements below then apply conditional on that context; they do not imply that the entire document is a first-order chain.

The label \ds{} denotes a standalone adaptation of the parallel-backbone architecture; we distinguish DSpark-Markov and DSpark-RNN by their sampling heads. DSpark-native denotes the original target-assisted decoding system. Both heads and their backbones are trained from scratch with fixed START/mask inputs; target features and verification are removed. Thus the training and sampling interfaces coincide, without removing expected inputs from a pretrained drafter. Section~\ref{sec:settings} specifies the implementation.

\subsection{Fixed backbone and previous-token head}
Under the selected abstraction, a backbone computes
\begin{equation}
    (U_1(C),\ldots,U_B(C))=f_\theta(C,\mathrm{BOS},\mathrm{MASK}_{1:B-1}),
    \label{eq:backbone}
\end{equation}
where $U_i$ denotes a vector of base logits. For fixed $C$, these vectors are the same across sampling runs, although they can differ across positions. The first token is drawn from $q_{\theta,1}(\cdot\mid C)$; subsequent tokens follow
\begin{equation}
    q_{\theta,i}(v\mid C,x_{<i})
    =\softmax\bigl(U_i(C)+T_\theta(x_{i-1},\cdot)\bigr)_v,
    \qquad i\geq2.
    \label{eq:local-head}
\end{equation}
The head may transform its available inputs nonlinearly; the dependence restriction is unchanged as long as it receives no additional sample-specific information. The additive form in Equation~\ref{eq:local-head} makes the motivating architecture concrete.

\begin{assumption}[Information available during one block]
\label{ass:interface}
The backbone inputs are fixed before sampling and its outputs are deterministic. The backbone is not updated using sampled tokens within the block. Each output conditional depends only on $C$, position-specific fixed features, and the previous sampled token. There is no additional shared latent variable, persistent head memory, or coupled sampling mechanism that carries earlier choices.
\end{assumption}

These are conditions on the complete generation procedure, not merely on the neural network diagram. A shared random seed is not itself a latent conditioning mechanism when it only implements the intended independent sampling draws; explicit reuse of a random variable to couple output choices would change the procedure. Figure~\ref{fig:information}(a) isolates one Markov sampling step: $U_i(C)$ remains fixed across runs with the same $C$, while the previous sampled token enters only the local head. In contrast, panel (b) shows a sample-dependent full-sequence state being updated across positions. The arrows specify information access, not a claim about model quality.

\begin{figure}[t]
\centering
\begin{tikzpicture}[x=1cm,y=1cm,>=Latex,
    box/.style={draw,rounded corners=2pt,minimum height=0.55cm,align=center,font=\small,inner sep=3pt},
    fixed/.style={box,draw=localblue,fill=stateblue},
    state/.style={box,draw=globalgreen,fill=stategreen},
    flow/.style={->,semithick},
    lane/.style={font=\scriptsize\bfseries,anchor=west,align=left}]
    \node[font=\small\bfseries,anchor=west] at (0,0.15)
        {(a) DSpark-Markov: one sampling step, $i\geq 2$};
    \node[lane,text=localblue] at (0,-0.65) {Fixed path};
    \node[fixed,minimum width=2.0cm] (context) at (3.2,-0.65) {$C$ + BOS / masks};
    \node[fixed,minimum width=2.25cm] (backbone) at (6.2,-0.65) {backbone $f_\theta$};
    \node[fixed,minimum width=1.8cm] (logits) at (9.2,-0.65) {$U_i(C)$};
    \draw[flow,localblue] (context) -- (backbone);
    \draw[flow,localblue] (backbone) -- (logits);

    \node[lane] at (0,-1.7) {Sampled path};
    \node[box,minimum width=1.4cm] (previous) at (6.2,-1.7) {$X_{i-1}$};
    \node[box,minimum width=2.35cm] (head) at (9.2,-1.7) {local head + sample};
    \node[box,minimum width=1.35cm] (current) at (12.1,-1.7) {$X_i$};
    \draw[flow] (previous) -- (head);
    \draw[flow,localblue] (logits) -- (head);
    \draw[flow] (head) -- (current);
    \node[font=\scriptsize,anchor=west] at (3.2,-2.32)
        {$q_{\theta,i}(\cdot\mid C,x_{<i})=q_{\theta,i}(\cdot\mid C,x_{i-1})$};

    \draw[black!25] (0,-2.78) -- (13.25,-2.78);
    \node[font=\small\bfseries,anchor=west] at (0,-3.13)
        {(b) Continuous generator: one state update, $s\to s+1$};
    \node[lane,text=globalgreen] at (0,-3.93) {Sampled state};
    \node[state,minimum width=2.0cm] (before) at (3.2,-3.93) {$Z_{1:B}^{(s)}$};
    \node[state,minimum width=3.0cm] (update) at (7.1,-3.93) {cross-position update $F_\theta$};
    \node[state,minimum width=2.0cm] (after) at (11.2,-3.93) {$Z_{1:B}^{(s+1)}$};
    \draw[flow,globalgreen] (before) -- (update);
    \draw[flow,globalgreen] (update) -- (after);
    \node[font=\scriptsize,anchor=west] at (3.2,-4.6)
        {Start from sample-specific $Z^{(0)}$; after $S$ updates, decode $X_{1:B}=D_\theta(Z^{(S)})$.};
\end{tikzpicture}
\caption{\textbf{Which information can change during generation?} (a) For fixed context $C$, the backbone produces the same position-specific logits $U_i(C)$ on every run. At step $i\geq2$, only the previous sampled token $X_{i-1}$ joins $U_i(C)$ at the Markov head; sampled tokens do not feed back into the backbone. The first token is sampled separately from $q_{\theta,1}$. DSpark-RNN adds persistent head memory and is not shown. (b) A continuous generator instead updates a sample-dependent full-sequence state, allowing information to pass across positions before decoding. Arrows show information access, not performance; $F_\theta$ abstracts a state update rather than prescribing one sampler.}
\label{fig:information}
\end{figure}

\subsection{Recurrent head with sampled-history memory}
\label{sec:rnn-interface}
For the recurrent variant, let $h_i(C)$ be the fixed backbone hidden state and $e_\theta$ a token embedding. A compact abstraction of the joint state and bias computation is
\begin{align}
    (s_i,b_i)&=G_\theta(s_{i-1},e_\theta(x_{i-1}),h_i(C)),\qquad s_0=0,\nonumber\\
    q^{\mathrm{RNN}}_{\theta,i}(v\mid C,x_{<i})
    &=\softmax\bigl(U_i(C)+b_i\bigr)_v.
    \label{eq:rnn-head}
\end{align}
Here $x_0$ is the fixed BOS/anchor input. The original DSpark head computes the next state and transition bias from the same concatenated input \citep{cheng2026dspark}; $G_\theta$ abstracts that computation, not a replacement cell. Even with fixed $h_i$ and $U_i$, $s_{i-1}$ can depend on earlier sampled tokens. This violates the no-memory condition in Assumption~\ref{ass:interface}. Recurrence removes that particular first-order restriction without guaranteeing successful learning or superior generation.

\subsection{Representation limit versus observed quality}
For a specified model class $\mathcal Q$, a representational limitation can be expressed by a positive approximation gap,
\begin{equation}
    \inf_{q\in\mathcal Q}\E_C\KL\bigl(p(X_{1:B}\mid C)\Vert q(X_{1:B}\mid C)\bigr)>0.
    \label{eq:ceiling}
\end{equation}
This is a representational claim, distinct from a fitted checkpoint's quality. Sequential token commitment alone does not imply a positive gap: the chain rule represents any joint distribution through full-history conditionals. Section~\ref{sec:structure} quantifies the cost of previous-token access for DSpark-Markov. Section~\ref{sec:continuous} then compares the fixed-backbone computation retained by a recurrent head with LangFlow's repeated updates of a sample-dependent full-sequence state.

\section{What Previous-Token Access Excludes}
\label{sec:structure}

DSpark-Markov conditions on the pre-block context $C$, but accesses choices made within the sampled block only through the preceding token. We quantify the best possible approximation error under this restriction, illustrate a positive gap with a delayed choice, and distinguish better probability fitting from access to more sampled history.

\subsection{The first-order restriction and its cost}
Under Assumption~\ref{ass:interface}, the generator belongs to
\begin{equation}
    \Qone=\left\{q:\ q(x_{1:B}\mid c)=q_1(x_1\mid c)
    \prod_{i=2}^{B}q_i(x_i\mid c,x_{i-1})\right\}.
    \label{eq:markov-class}
\end{equation}
This envelope permits arbitrary position- and context-dependent conditionals; the implemented low-rank head may impose additional restrictions. Membership implies
\begin{equation}
    X_i\perp X_{1:i-2}\mid(C,X_{i-1}),\qquad i\geq3.
    \label{eq:independence}
\end{equation}
Thus, two sampled prefixes with the same $C$ and preceding token must receive the same next-token distribution, even if their earlier choices differ. The missing information is earlier history that remains useful after conditioning on $C$ and the preceding token. Long-distance correlations conveyed by repeatedly copying that token are still possible.

\begin{proposition}[First-order conditional projection]
\label{prop:projection}
Let $p(C,X_{1:B})$ be a distribution over a finite token vocabulary, with $B\geq2$. Let $\Qone$ contain all conditional distributions in Equation~\ref{eq:markov-class}, including conditionals with zero probabilities. Then
\begin{equation}
    \inf_{q\in\Qone}\E_C\KL(p(\cdot\mid C)\Vert q(\cdot\mid C))
    =\sum_{i=3}^{B}I_p(X_i;X_{1:i-2}\mid C,X_{i-1}),
    \label{eq:projection}
\end{equation}
with an empty sum equal to zero. A minimizing distribution uses $q_1=p(X_1\mid C)$ and $q_i=p(X_i\mid C,X_{i-1})$ for $i\geq2$, on conditioning events of positive $p$-probability.
\end{proposition}

Matching conditional marginals and applying the entropy chain rule gives this standard projection identity (Appendix~\ref{app:proof}). Each mutual-information term measures what earlier tokens reveal about $X_i$ beyond $C$ and $X_{i-1}$. For an implemented subclass $\mathcal Q_\theta\subseteq\Qone$, the sum lower-bounds the best forward KL. Whenever it is positive, even ideal fitting cannot recover the reference distribution. A larger deterministic backbone or richer training targets cannot restore missing history while the sampling interface remains unchanged.

\subsection{A delayed-choice example}
\label{sec:witness}
For fixed context and $B\geq3$, draw $A\sim\operatorname{Bernoulli}(1/2)$ and define
\begin{equation}
    X_1=A,\qquad X_2=\cdots=X_{B-1}=\#,\qquad X_B=A.
    \label{eq:witness}
\end{equation}
Here $\#$ is a fixed separator; the two equally likely sequences are $0\,\#\cdots\#\,0$ and $1\,\#\cdots\#\,1$. At the last position, neither the preceding separator nor the fixed backbone features reveal the initial sampled choice.

The optimal first-order projection $q^\star$ retains the separators and independently selects the endpoints, assigning probability $1/4$ to each combination. Hence
\begin{equation}
    \KL(p\Vert q^\star)=\log 2,\qquad
    \Pr_{q^\star}[X_B\neq X_1]=\tfrac12.
    \label{eq:witness-values}
\end{equation}
These analytic values concern the optimal projection. Always selecting endpoint $0$ preserves consistency but drops a mode; encoding $A$ in a separator changes the target distribution. The total KL remains $\log 2$ as $B$ grows, so the example does not predict worsening per-token NLL or GenPPL with length.

\subsection{Better supervision versus more history}
The drafter can be strengthened in two different ways: fit probabilities better within the same interface, or expose more sampled history. Teacher distribution matching addresses the first. For distributions $p_T$ and $q$ on the same prefix, maximal one-token speculative acceptance is
\begin{equation}
    a(p_T,q)=1-\operatorname{TV}(p_T,q)
            =1-\tfrac12\|p_T-q\|_1.
    \label{eq:l1-acceptance}
\end{equation}
This motivates local L1 matching \citep{cheng2026dspark}, but richer targets do not expose missing sampled history. Appendix~\ref{app:l1-projection} shows how CE and L1 select different projections within the same restricted interface.

Recurrence addresses the second. DSpark-RNN (Section~\ref{sec:rnn-interface}) can carry the initial binary choice through a hidden state and need not belong to $\Qone$. This removes the specific first-order obstruction, while learning useful history remains a training problem. The backbone still stays fixed: adding head memory does not introduce repeated full-sequence updates. Section~\ref{sec:continuous} therefore examines a second distinction, the computation available after history access has been extended.

\Needspace*{12\baselineskip}
\section{Beyond Recurrent Memory: Full-Sequence Iteration}
\label{sec:continuous}

Section~\ref{sec:structure} identifies a restriction of the previous-token head. DSpark-RNN can escape that restriction and improves the evaluated drafter, yet a gap to LangFlow remains at lengths 128 and 1024. We now compare how the generators use sample-dependent computation: the recurrent head updates a prefix state, whereas LangFlow repeatedly updates a full-sequence state before token readout.

\subsection{A recurrent head leaves backbone features fixed}
In DSpark-RNN, $s_i$ evolves with the sampled prefix, while $h_i(C)$ and $U_i(C)$ remain fixed. With unconditional input $C=c_0$ and fixed $B$, these backbone outputs are also identical across sampling runs: they provide position-dependent features and base logits, while the recurrent head carries realized sample choices. Computation advances through the head one position at a time; later choices do not revise earlier backbone representations or committed tokens. Re-evaluating the deterministic backbone on unchanged inputs returns identical features. Backbone refinement requires feeding sample-dependent state back into it.

\subsection{Iterative updates before discrete readout}
A continuous generator instead maintains $Z^{(s)}\in\mathbb R^{B\times d}$, initialized from noise and transformed before token readout:
\begin{equation}
    Z^{(0)}\sim\pi_0,\qquad
    Z^{(s+1)}=F_{\theta,s}(Z^{(s)},C,\xi_s),\qquad
    X_{1:B}=D_\theta(Z^{(S)},C).
    \label{eq:continuous-state}
\end{equation}
Here $\xi_s$ denotes additional sampling noise when used. Each run starts with its own $Z^{(0)}$; each update processes the current full-sequence state, and discrete tokens are read out after $S$ updates. A shared sample-dependent state can coordinate the delayed-copy choices (Appendix~\ref{app:latent}). LangFlow also feeds previous predictions into subsequent evaluations through self-conditioning \citep{strudel2022sed,chen2026langflow}. Positions can therefore incorporate information developed elsewhere before final discrete readout.

Earlier positions can respond to later provisional states, and those responses can inform subsequent updates at the later positions. This repeated exchange occurs before discrete commitment; carrying a prefix summary through the output head does not itself supply this backbone feedback loop.

At fixed weights, NFE changes integration and self-conditioning trajectories. Section~\ref{sec:block-results} shows that additional denoiser evaluations improve LangFlow's GenPPL and MAUVE and change the length-128 ordering. The Markov projection in Section~\ref{sec:structure} characterizes a restricted distribution class; the NFE study demonstrates an additional improvement route through iterative generation.

\section{Experiments}
\label{sec:empirical}

\subsection{Experimental setup}
\label{sec:settings}
\paragraph{Task and models.}
We study unconditional OpenWebText generation at lengths $L\in\{16,128,1024\}$. A separate model is trained for each length and generates one block, so $B=L$. All models share the Qwen3 tokenizer and document BOS/EOS packing with random crops. DSpark-Markov and DSpark-RNN are trained from scratch on fixed START/mask inputs, without target features or verification, using token cross-entropy (CE). DSpark-Markov+L1 adds supervision from a frozen autoregressive teacher. Our LangFlow implementation uses embedding-space Gaussian diffusion, a learned noise schedule, and self-conditioning \citep{chen2026langflow}.

\paragraph{Training and sampling.}
All student backbones have width 768, 12 layers, 12 attention heads, and FFN width 3072; both sequential heads have rank 256. We compare EMA checkpoints at 100,000 updates, with global batch size 512, AdamW, BF16, peak learning rate $3\times10^{-4}$, 5,000 warmup steps, and a 250,000-step cosine horizon. EMA decay is 0.999. The length-1024 LangFlow checkpoint is taken at 100,000 updates from a longer training run and evaluated under the same protocol. DSpark variants use one backbone evaluation followed by categorical head sampling at temperature 1, without top-$k$ or top-$p$ truncation. LangFlow uses an Euler--EDM ODE sampler with self-conditioning and terminal argmax, with $L$ denoiser evaluations in the main comparison. All methods share the output-support mask. Appendix~\ref{app:record} gives implementation details.

\paragraph{Evaluation.}
With a frozen GPT-2 Large evaluator $r$, clean-text GenPPL is
\begin{equation}
    \operatorname{GenPPL}_r(\mathcal S)=
    \exp\left(-\frac{\sum_{x\in\mathcal S}\sum_{j=2}^{m_x}\log r(y^x_j\mid y^x_{<j})}{\sum_{x\in\mathcal S}(m_x-1)}\right),
    \label{eq:genppl}
\end{equation}
where $y^x$ is evaluator-tokenized text after removing special tokens, and $m_x$ is its scoring-window length. GenPPL uses 1,024 generated samples. MAUVE \citep{pillutla2021mauve} uses 5,120 generated and 5,120 fixed training-distribution reference samples per length, with 512 buckets. Both main evaluators have a 1,024-token window. The training seed is 42; main-comparison generations use their sample IDs as seeds. Figures report point estimates; Appendix~\ref{app:evaluation-checks} adds full-text scoring and repeated-evaluation checks at fixed checkpoints.

\subsection{The standalone generative gap}
\label{sec:main-results}
\begin{figure}[!htbp]
\centering
\includegraphics[width=0.9\linewidth]{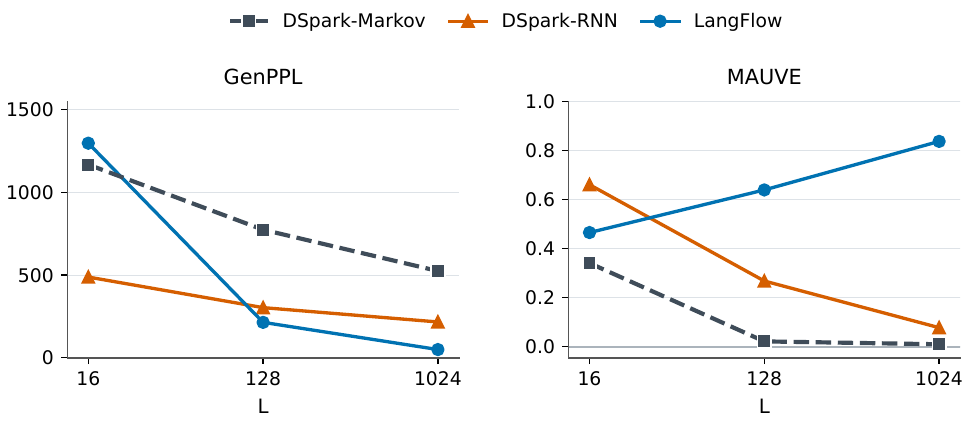}
\caption{\textbf{Standalone generation with Markov, recurrent, and diffusion models.} Raw metrics at the 100,000-update EMA checkpoint. Lower GenPPL and higher MAUVE are better. Each length uses separately trained models, with LangFlow NFE $=L$; lines guide the eye across lengths. RNN improves on Markov at every length and leads at length 16; LangFlow leads both drafters at lengths 128 and 1024.}
\label{fig:main-quality}
\end{figure}

We compare DSpark-Markov, DSpark-RNN, and LangFlow together to test whether extending the drafter's sampling history closes the generative gap. Figure~\ref{fig:main-quality} reports their raw GenPPL and MAUVE; Table~\ref{tab:main-results} gives the complete values. Both drafters use CE training, and LangFlow uses its default NFE $=L$.

\paragraph{Recurrence narrows the longer-sequence gap.}
At $L=128$, replacing the Markov head with the recurrent head improves GenPPL from 774.39 to 303.52 and MAUVE from 0.0221 to 0.2682, while LangFlow reaches 214.60 and 0.6392. At $L=1024$, recurrence improves the two scores from 524.79 and 0.0104 to 216.93 and 0.0780; LangFlow reaches 50.16 and 0.8370. Thus recurrence improves both metrics but leaves a gap at both longer lengths. This strengthening attempt expands history access while retaining a single fixed-backbone evaluation, as analyzed in Sections~\ref{sec:structure} and~\ref{sec:continuous}. Illustrative outputs at both lengths appear in Appendices~\ref{app:samples-main} and~\ref{app:samples-long}.

Full-text scoring preserves the $L=1024$ GenPPL ordering: Markov, RNN, and LangFlow reach 523.26, 215.69, and 50.15, respectively. Across 20 MAUVE clustering seeds, recurrence improves on Markov at every length, and the RNN--LangFlow ordering remains unchanged, including RNN's short-sequence advantage (Appendix~\ref{app:evaluation-checks}).

\paragraph{Short sequences favor the recurrent drafter.}
At $L=16$, DSpark-RNN leads with GenPPL 488.28 and MAUVE 0.6606. LangFlow has higher MAUVE than Markov (0.4655 versus 0.3409), but higher GenPPL (1297.25 versus 1167.77). The RNN--LangFlow ordering at lengths 16 and 128 holds at every recorded checkpoint from 10,000 to 100,000 updates (Appendix~\ref{app:checkpoints}). Comparisons are within each length: training-token exposure, evaluation context, and default denoising budget change across lengths. Appendix~\ref{app:samples-short} gives short-sequence examples.

The quality gains do not uniformly improve repetition. At $L=1024$, LangFlow repeats 7.33\% of source-token 4-grams, compared with 0.24\% for DSpark-RNN (Appendix~\ref{app:lexical}).

\subsection{The effect of teacher L1 supervision}
\label{sec:teacher-results}
We separately test whether teacher distribution matching improves the Markov drafter, holding its architecture and student update budget fixed. DSpark-Markov+L1 minimizes $0.1\mathcal L_{\mathrm{CE}}+0.9\mathcal L_1$, matching a frozen autoregressive Transformer's next-token probabilities. The teacher is trained separately at length 1024 and sees only the current student crop's available prefix. L1 sums absolute probability differences over the vocabulary and averages over positions and examples. We omit the original drafter's confidence loss and position weighting; the teacher is used only in training.

\begin{figure}[!htbp]
\centering
\includegraphics[width=0.9\linewidth]{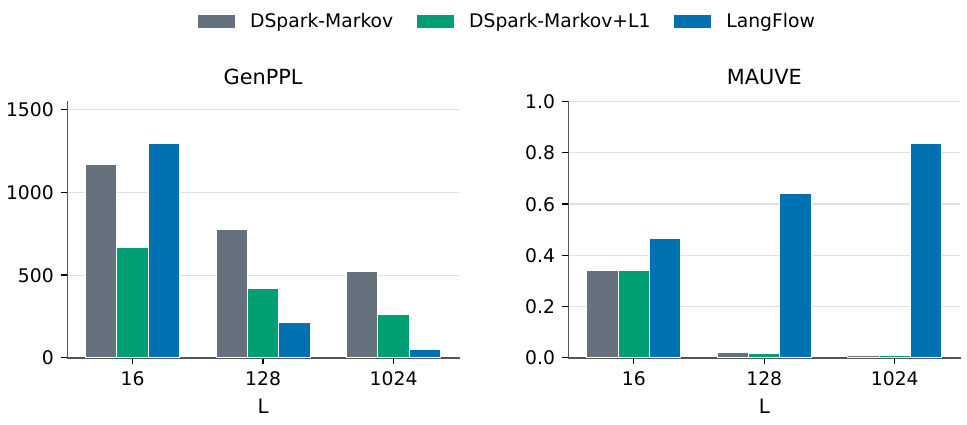}
\caption{\textbf{Teacher L1 supervision and the remaining gap to LangFlow.} Markov and Markov+L1 share the student architecture and update budget; LangFlow uses NFE $=L$. Lower GenPPL and higher MAUVE are better. L1 lowers GenPPL at every length without an observed MAUVE gain. LangFlow leads both Markov variants on both metrics at lengths 128 and 1024; length 16 has a mixed ordering. Bars are grouped by length with zero baselines; each length uses separately trained models.}
\label{fig:l1-supervision}
\end{figure}

Figure~\ref{fig:l1-supervision} compares Markov with and without L1, alongside LangFlow at its default budget. Supervision lowers GenPPL at all three lengths without an observed MAUVE gain. Markov+L1 reaches 419.82 and 0.0179 at $L=128$, and 262.55 and 0.0092 at $L=1024$, remaining behind LangFlow on both metrics. At $L=16$, Markov+L1 has lower GenPPL than LangFlow (668.48 versus 1297.25) but lower MAUVE (0.3404 versus 0.4655). Twenty independently generated batches per Markov variant at $L=16$ confirm the GenPPL benefit, with means 1152.99 and 677.18; the MAUVE difference is $-0.0078$ with a 95\% Welch interval of $[-0.0168,0.0013]$, showing no reliably detected gain (Appendix~\ref{app:evaluation-checks}). At $L=1024$, L1 also lowers corpus unigram entropy from 7.648 to 6.840 nats. Its improved evaluator predictability thus coexists with greater lexical concentration. L1 changes the fitting objective while leaving the Markov head's history access unchanged (Section~\ref{sec:structure}); Appendix~\ref{app:l1-projection} illustrates the associated projection effect. Text examples appear in Appendix~\ref{app:samples-l1}.

\subsection{Iteration changes the model ordering}
\label{sec:block-results}
\begin{figure}[!htbp]
\centering
\includegraphics[width=0.9\linewidth]{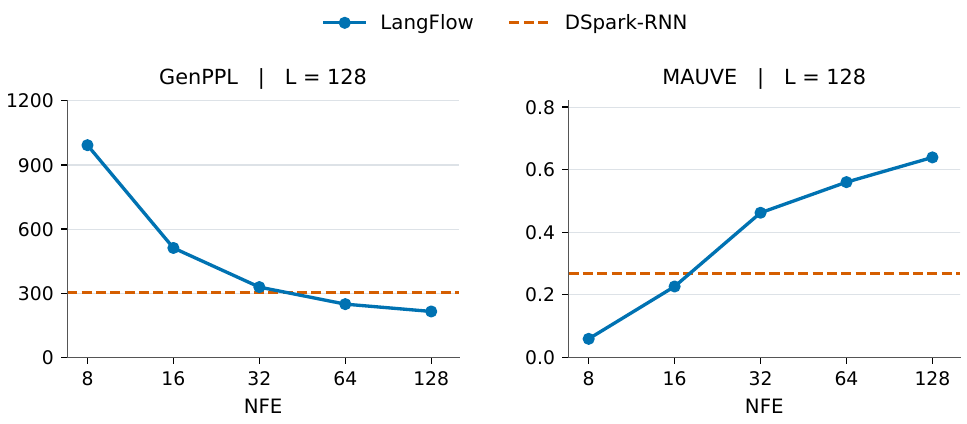}
\caption{\textbf{At length 128, additional iterations overtake DSpark-RNN.} Blue curves show LangFlow's raw GenPPL and MAUVE at fixed weights; orange dashed lines show DSpark-RNN. The first tested lead occurs at NFE64 for GenPPL and NFE32 for MAUVE. Thus, NFE64 is the first tested budget with a lead on both. Lower GenPPL and higher MAUVE are better. Values are point estimates; lines guide the eye.}
\label{fig:nfe-quality}
\end{figure}

The main comparison uses LangFlow's default NFE $=L$. To examine the effect of denoising steps separately, we hold each length's 100,000-update EMA checkpoint fixed and evaluate NFE $\in\{8,16,32,64,128\}$ for $L=16,128$. Every setting generates 5,120 samples, with 1,024 used for GenPPL and all 5,120 for MAUVE. Figure~\ref{fig:nfe-quality} plots raw GenPPL and MAUVE at $L=128$ against DSpark-RNN, the strongest evaluated drafter on both metrics. The short-sequence sweep appears in Figure~\ref{fig:nfe-short-quality}, and Table~\ref{tab:nfe-sweep} reports all values.

Every NFE increase on the tested grid improves both quality metrics at both lengths. At $L=128$, 32-step LangFlow reaches GenPPL 328.63 and MAUVE 0.4622, exceeding DSpark-RNN on MAUVE but having higher GenPPL. At 64 steps, both metrics improve on RNN (248.90 and 0.5601 versus 303.52 and 0.2682), making 64 the first tested budget to outperform all three DSpark variants on both metrics. Increasing to 128 steps further improves the scores to 214.60 and 0.6392. Appendix~\ref{app:samples-nfe} shows final-output examples across all five budgets.

At $L=16$, the best tested LangFlow sampler reaches GenPPL 731.62 and MAUVE 0.5787 at 128 steps; DSpark-RNN retains better values (488.28 and 0.6606). Iteration improves generation at both lengths, with a remaining short-sequence advantage for the recurrent drafter.


\section{Discussion}
\label{sec:discussion}

\paragraph{Architecture, supervision, and iteration.}
Recurrence strengthens the evaluated drafter, yet a gap to LangFlow remains at lengths 128 and 1024; RNN leads at length 16. The CE--L1 mixture lowers GenPPL without an observed MAUVE gain, while additional denoising at fixed weights changes the length-128 ordering.

\paragraph{Two routes to stronger generation.}
History access, revisable state, and target correction define complementary design axes, illustrated by causal proposal refinement \citep{wang2026xpress} and decoder-interface studies \citep{shen2026codar,du2026interface}. Our Markov bound addresses restricted history; RNN extends it with a fixed backbone, while LangFlow's fixed-weight NFE gains demonstrate an additional refinement route.

\section{Conclusion}
\label{sec:conclusion}
Recurrent augmentation improves DSpark-Markov at every tested length but leaves a GenPPL/MAUVE gap to LangFlow at lengths 128 and 1024 under its default budgets. RNN leads at length 16. The CE--L1 mixture lowers Markov's GenPPL without an observed MAUVE gain. At length 128, additional denoising at fixed weights improves both metrics and overtakes RNN from 64 tested steps onward. The Markov analysis identifies an information-access limit; the NFE results demonstrate iterative-computation gains. These findings establish a remaining gap after recurrence and a further generation improvement route for the evaluated LangFlow.

\label{main-text-end}

\clearpage
\bibliography{references}
\bibliographystyle{plainnat}
\clearpage
\appendix
\section{Derivation of the Conditional Projection Identity}
\label{app:proof}

\subsection{Notation and zero-probability events}
All logarithms are natural. Conditional entropies and mutual information are calculated under $p$. Conditional distributions on events of zero $p$-probability may be assigned arbitrarily. The class $\Qone$ permits zeros, so the minimizing conditional marginals are included. If a parameterization uses strictly positive softmax probabilities, the corresponding statement is an infimum in the closure; there need not be a finite parameter value attaining a boundary distribution.

\begin{table}[htbp]
\centering
\caption{Notation used in the architectural analysis.}
\begin{tabularx}{\linewidth}{@{}lY@{}}
\toprule
Symbol & Meaning \\
\midrule
$C$ & All pre-block information available to the generator. \\
$c_0$ & Fixed conditioning input for unconditional, single-block generation. \\
$B$, $L$ & Generation block size and total output length, respectively. \\
$p$, $q$ & Reference and generator distributions over token blocks. \\
$U_i(C)$ & Fixed base-logit vector at position $i$. \\
$\Qone$ & All position-dependent first-order conditional block distributions. \\
$Z^{(s)}$ & Sample-dependent continuous state after update $s$. \\
$r$ & Frozen external evaluator, distinct from $p$ and $q$. \\
\bottomrule
\end{tabularx}
\end{table}

\subsection{Proof of Proposition~\ref{prop:projection}}
For any $q\in\Qone$, expand its factorization inside the expected forward KL:
\begin{align}
    \E_C\KL(p(\cdot\mid C)\Vert q(\cdot\mid C))
    ={}&-H_p(X_{1:B}\mid C)-\E_p\log q_1(X_1\mid C)\nonumber\\
    &-\sum_{i=2}^B\E_p\log q_i(X_i\mid C,X_{i-1}).
    \label{eq:proof-start}
\end{align}
Each cross-entropy term has the familiar conditional decomposition
\begin{align}
    -\E_p\log q_1(X_1\mid C)
    ={}&H_p(X_1\mid C)
       +\E_C\KL\bigl(p(X_1\mid C)\Vert q_1(X_1\mid C)\bigr),\\
    -\E_p\log q_i(X_i\mid C,X_{i-1})
    ={}&H_p(X_i\mid C,X_{i-1})\nonumber\\
    &+\E_{C,X_{i-1}}\KL\bigl(p(X_i\mid C,X_{i-1})\Vert q_i(X_i\mid C,X_{i-1})\bigr).
\end{align}
Substituting and expanding the joint entropy by the chain rule gives
\begin{align}
    \E_C\KL(p\Vert q)
    ={}&\sum_{i=3}^B I_p(X_i;X_{1:i-2}\mid C,X_{i-1})\nonumber\\
    &+\E_C\KL\bigl(p(X_1\mid C)\Vert q_1(X_1\mid C)\bigr)\nonumber\\
    &+\sum_{i=2}^B\E_{C,X_{i-1}}\KL\bigl(p(X_i\mid C,X_{i-1})\Vert q_i(X_i\mid C,X_{i-1})\bigr).
    \label{eq:full-decomposition}
\end{align}
All remaining KL terms are nonnegative and vanish simultaneously by matching the indicated conditional marginals. This proves the identity. Any smaller architecture class has an infimum at least as large as the infimum over $\Qone$. \hfill$\square$

The result characterizes the distributional envelope, not a particular learning algorithm. Replacing the fitting objective can change which $q$ is learned without altering the first term of Equation~\ref{eq:full-decomposition}. Conversely, changing the information interface can change the envelope itself, which is why memory and latent-state modifications require separate analysis.

\section{Exact Analysis of the Delayed-Copy Distribution}
\label{app:witness}
For the distribution in Equation~\ref{eq:witness}, the only nonzero conditional mutual information in Equation~\ref{eq:projection} is the final one:
\begin{equation}
    I_p(X_B;X_{1:B-2}\mid X_{B-1})
    =H_p(X_B\mid X_{B-1})-H_p(X_B\mid X_{1:B-1})
    =\log 2.
\end{equation}
All intermediate tokens are deterministic separators. Matching the one-step conditionals yields the probabilities in Table~\ref{tab:analytic}. Both valid sequences have $p(x)=1/2$ and $q^\star(x)=1/4$, hence
\begin{equation}
    \KL(p\Vert q^\star)=2\left(\tfrac12\log\frac{1/2}{1/4}\right)=\log2.
\end{equation}

\begin{table}[htbp]
\centering
\caption{\textbf{Analytic probabilities, not experimental results.} Every interior token is $\#$. The optimal first-order projection preserves endpoint marginals but introduces two inconsistent combinations.}
\label{tab:analytic}
\begin{tabular}{@{}cccc@{}}
\toprule
$(X_1,X_B)$ & Reference $p$ & Projection $q^\star$ & Endpoint consistency \\
\midrule
$(0,0)$ & $1/2$ & $1/4$ & Yes \\
$(0,1)$ & $0$   & $1/4$ & No \\
$(1,0)$ & $0$   & $1/4$ & No \\
$(1,1)$ & $1/2$ & $1/4$ & Yes \\
\bottomrule
\end{tabular}
\end{table}

The $1/2$ mismatch rate belongs to the forward-KL minimizer, not to every first-order generator. A generator that chooses only one endpoint value has zero mismatch but misses a target mode; assigning zero probability to that mode makes its forward KL infinite. This distinction prevents consistency alone from being mistaken for distribution matching. The constant total gap also makes clear that increasing $B$ does not force a worsening per-token score.

\section{A Shared Continuous Latent Escapes the Toy Obstruction}
\label{app:latent}
Draw $U\sim\mathcal N(0,1)$ and let $A=\mathbf1\{U\geq0\}$. Define a continuous block state whose first and last positions retain $U$, while intermediate positions carry a fixed separator representation. A position-aware readout maps the signs of the endpoint states to $0$ or $1$ and emits $\#$ elsewhere. The resulting token distribution is exactly the reference distribution in Equation~\ref{eq:witness}.

The construction illustrates a dependency pathway, not an implementation of LangFlow. In particular, a map that explicitly copies a latent need not be an invertible finite-time flow, and threshold readout is not an assertion about finite softmax parameters. Demonstrating a strict class inclusion for a particular CDLM would require specifying its state dynamics, capacity, readout, and approximation criterion. We make no such inclusion claim here. The construction only shows why a shared sample-dependent state is not bound by the previous-token interface.

\section{Schematic Sampling Procedures}
\label{app:sampling}
The procedures below summarize the three sampling interfaces. Appendix~\ref{app:record} specifies their experimental instantiations, including the recurrent cell, initialization, and continuous sampler.

\paragraph{Fixed backbone with a first-order head.}
\begin{enumerate}
    \item Fix pre-block information $C$ and generation length $B$.
    \item Compute $U_{1:B}$ once from the fixed backbone input.
    \item Draw $X_1\sim q_1(\cdot\mid C)$.
    \item For $i=2,\ldots,B$, draw $X_i\sim q_i(\cdot\mid C,X_{i-1})$ without updating $U_{1:B}$ or a recurrent state.
    \item Return $X_{1:B}$ without target verification.
\end{enumerate}

\paragraph{Fixed backbone with a recurrent head.}
\begin{enumerate}
    \item Fix $C$, $B$, and $x_0$; initialize $s_0=0$ at the block boundary.
    \item Compute $h_{1:B}$ and $U_{1:B}$ once from the fixed backbone input.
    \item For $i=1,\ldots,B$, compute $(s_i,b_i)$ as in Equation~\ref{eq:rnn-head}, sample $X_i$ from the resulting logits, and retain $s_i$ and $X_i$ for the next step.
    \item Return $X_{1:B}$ without target verification.
\end{enumerate}

\paragraph{Continuous state with cross-position updates.}
\begin{enumerate}
    \item Fix $C$, $B$, and the method's sampling configuration.
    \item Draw the initial state $Z^{(0)}\sim\pi_0$.
    \item Apply the method-specific updates to the whole state; include self-conditioning and stochastic updates when the method uses them.
    \item Decode the final state into $X_{1:B}$ using the specified readout.
    \item Return $X_{1:B}$ without target verification.
\end{enumerate}
All output laws must be assessed under the procedures actually used. For example, whole-sequence rejection, reranking, or a shared random variable in the first procedure can alter its output class. Such modifications are part of the generator, not invisible evaluation details.

\section{An Illustrative CE--L1 Projection Difference}
\label{app:l1-projection}
Consider an unconditional three-token distribution $X_1=A$, $X_2=\#$, $X_3=A$, with $A\sim\operatorname{Bernoulli}(a)$ and $0<a<1/2$. An exact teacher sees $A$ at the last position; a first-order student sees only the separator and predicts $q(X_3=1\mid\#)=t$. The expected final-position objectives are
\begin{equation}
    \mathcal L_{\mathrm{CE}}(t)=-a\log t-(1-a)\log(1-t),
    \qquad
    \mathcal L_1(t)=2\bigl[a(1-t)+(1-a)t\bigr].
    \label{eq:toy-l1-projection}
\end{equation}
CE is minimized at $t=a$, preserving the marginal frequency. L1 is minimized at $t=0$, favoring the majority outcome when the distinguishing history is unavailable. For $a=1/4$, the mixture $0.1\mathcal L_{\mathrm{CE}}+0.9\mathcal L_1$ has the unique minimizer
\begin{equation}
    t^\star=\frac{1-\sqrt{0.91}}{1.8}\simeq0.0256.
    \label{eq:toy-l1-optimum}
\end{equation}
This follows by setting its derivative to zero: $t-0.025-0.9t^2=0$, taking the root in $(0,1)$. Neither student reconstructs the joint dependency, but the objectives favor different marginal behavior. An adequate recurrent student can instead condition on $A$ and minimize both objectives simultaneously. This is an analytic counterexample to objective equivalence under a restricted interface, not a claim that L1 always collapses modes, that our teacher is exact, or that the example predicts GenPPL/MAUVE.

\section{Experimental Details}
\label{app:record}

\subsection{Data representation and output support}
OpenWebText documents are tokenized with the Qwen3 vocabulary, bracketed by document BOS/EOS markers, packed into streams, and randomly cropped to the training length. The shuffle buffer contains 10,000 examples. The model vocabulary allocates 151,936 entries, of which 151,669 belong to the tokenizer. The fixed scaffold uses START token 151644 and MASK token 151660; document boundary tokens are 151643 and 151645. Generation excludes START, MASK, and IDs outside the tokenizer vocabulary for every model, including at each LangFlow denoiser evaluation. Document boundaries remain allowed. Samples have exactly $L$ generated tokens, without early stopping at EOS. The control scaffold is distinct from the observed text prefix used for teacher forcing.

\subsection{Standalone sampling heads}
The DSpark backbone applies bidirectional attention to $[\mathrm{START},\mathrm{MASK},\ldots,\mathrm{MASK}]$. Its features $h_i\in\mathbb R^{768}$ and base logits $U_i$ are computed once per sample. Both heads use a token table $E\in\mathbb R^{151936\times256}$ and a vocabulary projection $W\in\mathbb R^{151936\times256}$. DSpark-Markov adds $W E[x_{i-1}]$ to $U_i$. DSpark-RNN uses a 256-dimensional state and a joint affine map:
\begin{align}
    z_i&=[s_{i-1};E[x_{i-1}];h_i],\qquad
    (g_i,a_i,o_i)=Jz_i+b,\nonumber\\
    s_i&=\sigma(g_i)\odot s_{i-1}+
    (1-\sigma(g_i))\odot\tanh(a_i),\qquad
    b_i=W\tanh(o_i).
    \label{eq:implemented-rnn}
\end{align}
The input to $J$ has dimension $2\times256+768$ and its output has dimension $3\times256$. The transition bias $b_i$ and next state $s_i$ are computed from the same input; the bias does not require a separate transformation of the new state. The state starts at zero, persists throughout the sample, and resets only between samples. Training uses the shifted ground-truth sequence, with START at the first position, and backpropagates through the full recurrent sequence. Inference substitutes previously sampled tokens. No target features, verification, reranking, or rejection step is used.

\subsection{Continuous model and optimization}
LangFlow corrupts learned token embeddings with Gaussian noise and trains a denoising predictor with a learned noise schedule. The clean embedding scale is 0.1. The learnable Gumbel noise schedule is initialized with location 0.117829814, scale 0.852, target entropy 7.02, and quantile cutoff $10^{-5}$; this schedule does not change the Gaussian noise distribution itself. The training objective combines token reconstruction cross-entropy with schedule fitting. Self-conditioning is enabled with probability 0.25 during training and used during ODE sampling. The analytic logit-skip contribution warms up over 5,000 steps. Sampling initializes from Gaussian noise, performs the specified number of Euler--EDM updates, and takes the final token argmax.

Shared optimization settings are AdamW with $(\beta_1,\beta_2)=(0.9,0.999)$, weight decay 0.01, gradient clipping at norm 1, dropout 0.1, and BF16 computation. The base learning rate warms up to $3\times10^{-4}$ and follows a cosine schedule with a $3\times10^{-5}$ floor and 250,000-step horizon. The comparison uses the checkpoint at 100,000 steps, when the base learning rate is approximately $2.12\times10^{-4}$. The length-1024 LangFlow checkpoint is taken at that step from a run scheduled for 250,000 updates. All trainable student parameters share this optimizer schedule. Each student receives 51.2 million length-specific training examples, corresponding to 0.8192, 6.5536, and 52.4288 billion target-token positions at $L=16,128,1024$, respectively. These are exposure counts, not counts of distinct tokens. The main comparison uses the common 100,000-update checkpoint, without choosing a different step for each model.

\subsection{Teacher supervision}
The teacher is a causal Transformer with width 768, 12 layers, 12 attention heads, FFN width 3072, and the same tokenizer and data preprocessing. It is trained separately for 100,000 updates at length 1024; the frozen EMA is used for distillation. The teacher receives START followed by the observed prefix of the current student crop, without preceding text outside that crop or future target tokens. With $p_T$ its conditional probabilities, the student objective is
\begin{align}
    \mathcal L_{\mathrm{CE}}
    &=-\E_{x}\frac1L\sum_{i=1}^{L}\log q_{\theta,i}(x_i\mid x_{<i}),\nonumber\\
    \mathcal L_1
    &=\E_x\frac1L\sum_{i=1}^{L}\sum_{v\in\V}
    \left|p_T(v\mid x_{<i})-q_{\theta,i}(v\mid x_{<i})\right|,\nonumber\\
    \mathcal L_{\mathrm{distill}}
    &=0.1\mathcal L_{\mathrm{CE}}+0.9\mathcal L_1.
    \label{eq:distillation-objective}
\end{align}
The notation $x_{<i}$ specifies the supplied prefix; DSpark-Markov accesses it only through $x_{i-1}$. Probabilities are computed in FP32 at temperature 1 over the full model vocabulary. The sum over vocabulary entries is not divided by vocabulary size; its range is $[0,2]$. Position means exclude padding, and examples are equally weighted. No teacher gradient enters the student optimizer or EMA. Compared with CE-only training, this ablation changes both the presence of L1 and the CE coefficient. Its outcome is therefore the effect of the stated mixture, not an isolated estimate of one loss term at equal gradient scale.

\subsection{Parameter and computation accounting}
Table~\ref{tab:parameters} reports stored parameter counts and counts excluding unused scaffold-embedding rows. The standalone DSpark variants access only two rows of their $151936\times768$ backbone input table, leaving 116,685,312 stored parameters outside the generation path. This table is separate from the head's previous-token embedding, which remains active. The recurrent cell adds 983,808 parameters. Matching backbone geometry therefore does not exactly match the number of parameters participating in generation.

\begin{table}[htbp]
\centering
\small
\caption{Student parameter accounting. The last column subtracts only unused scaffold-embedding rows; it is not a measure of effective statistical capacity. The teacher is excluded from student counts.}
\label{tab:parameters}
\begin{tabular}{@{}lrr@{}}
\toprule
Model & Stored parameters & Excluding unused rows \\
\midrule
DSpark-Markov / DSpark-Markov+L1 & 396,316,800 & 279,631,488 \\
DSpark-RNN & 397,300,608 & 280,615,296 \\
LangFlow & 327,085,955 & 327,085,955 \\
\bottomrule
\end{tabular}
\end{table}

DSpark-Markov+L1 incurs separate teacher training and online teacher forward passes during student training. The teacher is absent from standalone inference. Sequential models use one full backbone pass followed by $L$ lightweight head steps; LangFlow uses one full denoiser pass per NFE.

\subsection{Metric computation and sample counts}
Special tokens are removed before text is re-tokenized for GPT-2 Large, which is distinct from the distillation teacher. GenPPL aggregates negative log-likelihood sums and scored-token counts across samples before exponentiation. The first evaluator token is unscored; no additional BOS target is inserted. The evaluator uses BF16 and a maximum window of 1,024 evaluator tokens, truncating longer tokenizations. MAUVE also uses GPT-2 Large features with a 1,024-token window, 512 buckets, and the same cleaning procedure. Consequently, the metrics describe the evaluator-visible text rather than guaranteeing full coverage of any sequence exceeding that window.

Within each length, all models use the same 5,120 reference texts, sampled with seed 12345 from the training distribution and cropped using the generation tokenizer. GenPPL uses generated sample IDs 0--1023; MAUVE uses IDs 0--5119. The fixed-checkpoint NFE sweep uses the same counts: 5,120 generated samples per setting, with the first 1,024 used for GenPPL and all 5,120 for MAUVE. The main comparison and NFE sweep report point estimates; Appendix~\ref{app:evaluation-checks} adds evaluation checks at fixed trained checkpoints. Unigram entropy is measured on generated tokenizer IDs over the corpus. It differs from the entropy of a model's terminal probability vector, which excludes randomness introduced earlier in diffusion sampling.

\section{Additional Quantitative Results}
\label{app:quantitative}

\subsection{Complete final-checkpoint comparison}
Table~\ref{tab:main-results} lists the raw values underlying Figures~\ref{fig:main-quality} and~\ref{fig:l1-supervision}. Every entry uses the 100,000-update EMA checkpoint; the length-1024 LangFlow entry is evaluated with 1024 denoising steps and the common scoring protocol. Comparisons use the same sample counts within each length.

\begin{table}[!htbp]
\centering
\small
\setlength{\tabcolsep}{4pt}
\caption{OpenWebText unconditional generation at the 100,000-update EMA checkpoint. Bold marks the best value within each length. LangFlow uses NFE $=L$; drafters use one backbone pass and $L$ head steps. GenPPL and MAUVE use 1,024 and 5,120 generated samples, respectively.}
\label{tab:main-results}
\begin{tabular}{@{}lrrrrrr@{}}
\toprule
& \multicolumn{2}{c}{$L=16$} & \multicolumn{2}{c}{$L=128$} & \multicolumn{2}{c}{$L=1024$} \\
\cmidrule(lr){2-3}\cmidrule(lr){4-5}\cmidrule(l){6-7}
Model & GenPPL$\downarrow$ & MAUVE$\uparrow$ & GenPPL$\downarrow$ & MAUVE$\uparrow$ & GenPPL$\downarrow$ & MAUVE$\uparrow$ \\
\midrule
DSpark-Markov & 1167.77 & 0.3409 & 774.39 & 0.0221 & 524.79 & 0.0104 \\
DSpark-RNN & \textbf{488.28} & \textbf{0.6606} & 303.52 & 0.2682 & 216.93 & 0.0780 \\
LangFlow & 1297.25 & 0.4655 & \textbf{214.60} & \textbf{0.6392} & \textbf{50.16} & \textbf{0.8370} \\
\midrule
DSpark-Markov+L1 & 668.48 & 0.3404 & 419.82 & 0.0179 & 262.55 & 0.0092 \\
\bottomrule
\end{tabular}
\end{table}

\subsection{Evaluation along training}
\label{app:checkpoints}
\begin{figure}[!htbp]
\centering
\includegraphics[width=\linewidth]{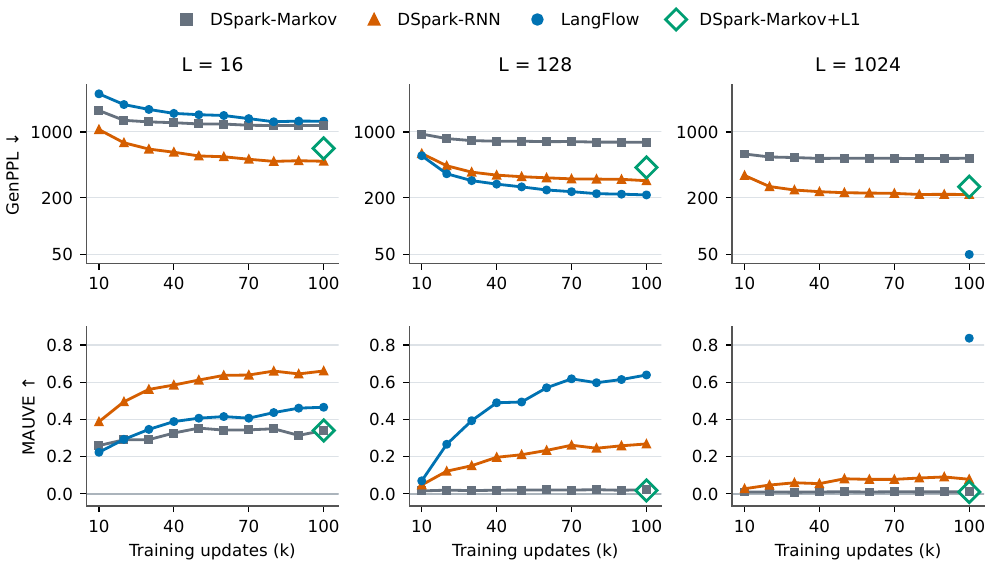}
\caption{Quality along training under the common evaluation protocol and default sampling budgets. All 84 recorded points are shown, sorted by update count. DSpark-Markov+L1 at every length and LangFlow at $L=1024$ have only a 100,000-update point under this protocol; no intermediate curve is inferred. Outlined L1 diamonds keep the nearly coincident Markov endpoints visible at their measured coordinates. The curves are checkpoint trajectories, not averages over training seeds.}
\label{fig:checkpoint-quality}
\end{figure}

Figure~\ref{fig:checkpoint-quality} shows the available evaluations from 10,000 to 100,000 updates at the default sampling budgets. At every recorded checkpoint, DSpark-RNN leads LangFlow on both metrics at $L=16$, and LangFlow leads DSpark-RNN on both metrics at $L=128$. The main comparison therefore does not select an exceptional checkpoint for either ordering. At $L=1024$, Markov MAUVE stays near 0.01 across the recorded points; RNN improves it but remains below the LangFlow endpoint. The common-protocol data contain no earlier LangFlow evaluations at this length.

\clearpage
\subsection{Generation quality across denoising steps}
\begin{figure}[!htbp]
\centering
\includegraphics[width=\linewidth]{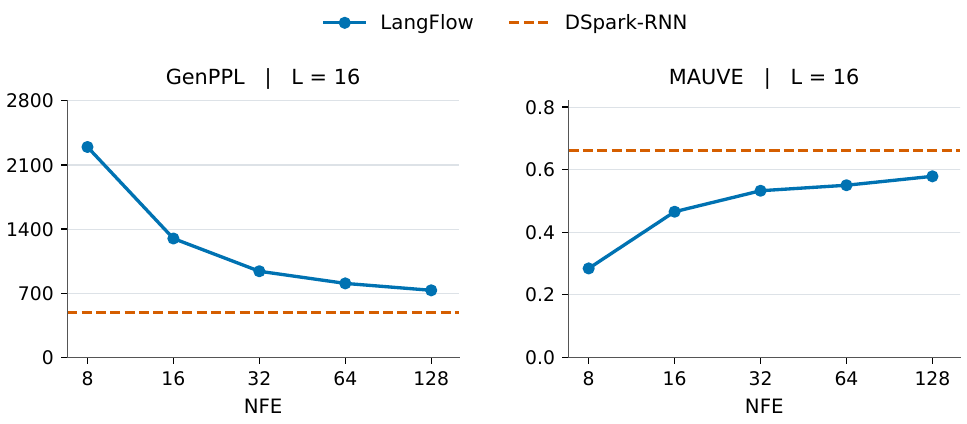}
\caption{\textbf{At length 16, iteration improves quality without overtaking DSpark-RNN.} Raw LangFlow GenPPL and MAUVE improve across all five tested budgets, while the recurrent drafter retains the better value on both metrics. At NFE128, LangFlow reaches GenPPL 731.62 and MAUVE 0.5787, versus RNN's 488.28 and 0.6606. Weights are fixed; values are point estimates, and connecting lines guide the eye.}
\label{fig:nfe-short-quality}
\end{figure}

Table~\ref{tab:nfe-sweep} gives the raw values underlying Figures~\ref{fig:nfe-quality} and~\ref{fig:nfe-short-quality}. Each length uses one fixed 100,000-update EMA checkpoint across all five NFE settings, so the comparison isolates changes to the sampling procedure from changes to the learned weights.

\begin{table}[!htbp]
\centering
\small
\setlength{\tabcolsep}{6pt}
\caption{Fixed-checkpoint LangFlow generation quality across denoising budgets. All settings use 1,024 GenPPL and 5,120 MAUVE samples; values are point estimates.}
\label{tab:nfe-sweep}
\begin{tabular}{@{}rrrrr@{}}
\toprule
& \multicolumn{2}{c}{$L=16$} & \multicolumn{2}{c}{$L=128$} \\
\cmidrule(lr){2-3}\cmidrule(l){4-5}
NFE & GenPPL$\downarrow$ & MAUVE$\uparrow$ & GenPPL$\downarrow$ & MAUVE$\uparrow$ \\
\midrule
8 & 2296.18 & 0.2842 & 992.48 & 0.0591 \\
16 & 1297.25 & 0.4655 & 511.70 & 0.2265 \\
32 & 939.76 & 0.5326 & 328.63 & 0.4622 \\
64 & 807.12 & 0.5502 & 248.90 & 0.5601 \\
128 & 731.62 & 0.5787 & 214.60 & 0.6392 \\
\bottomrule
\end{tabular}
\end{table}

\clearpage
\subsection{Lexical concentration and repetition}
\label{app:lexical}
\begin{figure}[!htbp]
\centering
\includegraphics[width=\linewidth]{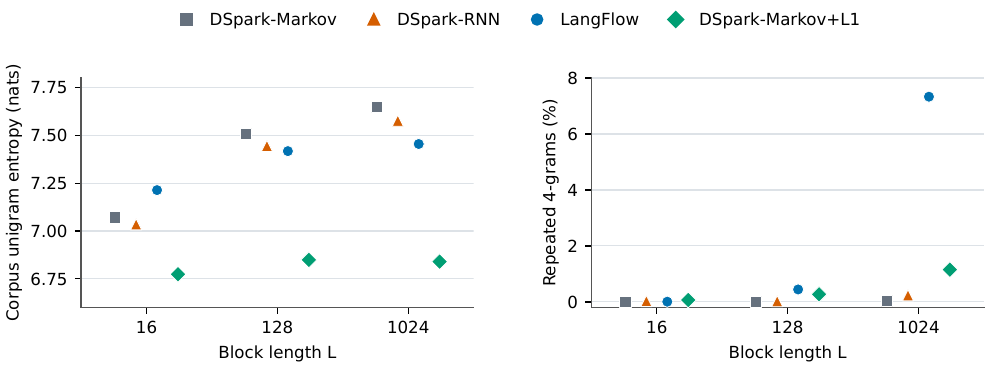}
\caption{Lexical diagnostics on 1,024 full source-token sequences per model and length. Entropy measures corpus-level lexical spread; repeated 4-grams measure within-sequence duplication. Neither is a standalone measure of text quality. Points at different lengths use different trained models and corpus token counts.}
\label{fig:lexical-diagnostics}
\end{figure}

Figure~\ref{fig:lexical-diagnostics} and Table~\ref{tab:lexical} complement the two main quality metrics. Corpus unigram entropy is $H=-\sum_v\hat p(v)\log\hat p(v)$ on the full generated tokenizer-ID corpus. Repeated 4-grams are counted within each sequence: occurrences after the first instance of a 4-gram are summed and divided by the total number of 4-gram positions. Both diagnostics use the first 1,024 generated samples, including the full length-1024 sequences, without evaluator re-tokenization or its window truncation.

Teacher supervision lowers corpus entropy relative to the Markov baseline at all three lengths. LangFlow's favorable GenPPL and MAUVE at length 1024 coexist with a higher repeated-4-gram fraction: 7.33\%, versus 0.03\% for Markov, 0.24\% for RNN, and 1.15\% for Markov+L1. Low repetition alone is also insufficient: the Markov baseline has low duplication but poor MAUVE at the two longer lengths. These diagnostics separate lexical concentration and local repetition from the distributional comparison.

\begin{table}[!htbp]
\centering
\small
\setlength{\tabcolsep}{4pt}
\caption{Corpus unigram entropy $H$ (nats) and repeated 4-grams $R_4$ (percent) on 1,024 full source-token sequences. Values are rounded independently.}
\label{tab:lexical}
\begin{tabular}{@{}lrrrrrr@{}}
\toprule
& \multicolumn{2}{c}{$L=16$} & \multicolumn{2}{c}{$L=128$} & \multicolumn{2}{c}{$L=1024$} \\
\cmidrule(lr){2-3}\cmidrule(lr){4-5}\cmidrule(l){6-7}
Model & $H$ & $R_4$ & $H$ & $R_4$ & $H$ & $R_4$ \\
\midrule
DSpark-Markov & 7.071 & 0.000 & 7.509 & 0.009 & 7.648 & 0.034 \\
DSpark-RNN & 7.035 & 0.030 & 7.445 & 0.029 & 7.577 & 0.237 \\
LangFlow & 7.214 & 0.008 & 7.418 & 0.445 & 7.455 & 7.332 \\
\midrule
DSpark-Markov+L1 & 6.774 & 0.068 & 6.849 & 0.270 & 6.840 & 1.153 \\
\bottomrule
\end{tabular}
\end{table}

\clearpage
\subsection{Evaluation checks for the generative gap}
\label{app:evaluation-checks}
All checks use the same 100,000-update EMA checkpoints as the main comparison.

\paragraph{Full-text GenPPL.}
At $L=1024$, we score the same 1,024 complete outputs with 1,024-token context windows and stride 512, counting each predictable target once. The original prefix scorer covers 96.03\% of LangFlow's targets; adding the remaining targets changes its GenPPL from 50.16 to 50.15. Table~\ref{tab:full-text} preserves the gap after recurrence and L1 supervision. Intervals use 10,000 complete-sample bootstrap resamples of token-weighted GenPPL.

\begin{table}[!htbp]
\centering
\small
\setlength{\tabcolsep}{5pt}
\caption{GenPPL at $L=1024$: original prefix and complete-output scoring.}
\label{tab:full-text}
\begin{tabular}{@{}lrrc@{}}
\toprule
Model & Prefix & Full text & 95\% bootstrap interval \\
\midrule
DSpark-Markov & 524.79 & 523.26 & $[519.88,526.66]$ \\
DSpark-RNN & 216.93 & 215.69 & $[212.27,219.17]$ \\
LangFlow & 50.16 & 50.15 & $[48.89,51.42]$ \\
\midrule
DSpark-Markov+L1 & 262.55 & 262.41 & $[260.49,264.24]$ \\
\bottomrule
\end{tabular}
\end{table}

\paragraph{MAUVE clustering sensitivity.}
We keep generated and reference texts fixed and refit MAUVE with clustering seeds 0--19, retaining 512 buckets. Table~\ref{tab:clustering} reports minimum and maximum score differences across these seeds. Recurrence improves on Markov at every length; LangFlow remains ahead of RNN at $L=128,1024$, while RNN leads at $L=16$. These ranges measure clustering sensitivity.

\begin{table}[!htbp]
\centering
\small
\setlength{\tabcolsep}{5pt}
\caption{MAUVE score-difference ranges over 20 clustering seeds.}
\label{tab:clustering}
\begin{tabular}{@{}rcc@{}}
\toprule
$L$ & RNN $-$ Markov & LangFlow $-$ RNN \\
\midrule
16 & $[0.2585,0.3298]$ & $[-0.2460,-0.1541]$ \\
128 & $[0.2161,0.2540]$ & $[0.3405,0.4295]$ \\
1024 & $[0.0561,0.0834]$ & $[0.7282,0.7987]$ \\
\bottomrule
\end{tabular}
\end{table}

\paragraph{Independent sampling of the supervision comparison.}
At $L=16$, each Markov variant generates 20 independent batches of 5,120 samples using disjoint sampling-seed streams and the original temperature and sampling procedure. GenPPL uses the first 1,024 samples of each batch; MAUVE uses all samples with the original reference set and clustering seed 0. Table~\ref{tab:independent-sampling} gives means of batch scores and approximate Welch intervals for their expected differences. L1 lowers GenPPL without a reliably detected MAUVE gain. The intervals are conditional on the checkpoints and evaluation settings.

\begin{table}[!htbp]
\centering
\small
\setlength{\tabcolsep}{5pt}
\caption{Mean scores from 20 independent $L=16$ batches per method. Differences are Markov+L1 minus Markov, with approximate 95\% Welch intervals.}
\label{tab:independent-sampling}
\begin{tabular}{@{}lrrrc@{}}
\toprule
Metric & Markov & Markov+L1 & Difference & 95\% interval \\
\midrule
GenPPL & 1152.99 & 677.18 & $-475.81$ & $[-491.95,-459.68]$ \\
MAUVE & 0.3424 & 0.3347 & $-0.0078$ & $[-0.0168,0.0013]$ \\
\bottomrule
\end{tabular}
\end{table}

\FloatBarrier

\clearpage
\section{Generated Samples and Text Organization}
\label{app:qualitative}

These unconditional outputs illustrate the text behavior accompanying the generative gap and the separate supervision and iteration comparisons. All use 100,000-update EMA checkpoints. Original English text is retained, including errors, incomplete sentences, and document-boundary tokens; whitespace is reflowed for typesetting. The same generations appear in both language versions. Selection is post hoc and illustrative: evaluator-NLL constraints do not imply random sampling or median semantic quality. Aggregate results establish the model ordering; individual outputs do not establish an architectural cause. Sample PPL below scores each complete clean text, using the full-text scorer at $L=1024$, and differs from token-weighted corpus GenPPL. MAUVE is a corpus metric and is not assigned to individual outputs.

\subsection{Length-128 comparison}
\label{app:samples-main}
For each model, we identify the 11 outputs nearest to its median per-sample mean evaluator NLL among the first 1,024 GenPPL samples, then select an example illustrating text organization. The selected DSpark-Markov, DSpark-RNN, and LangFlow outputs have NLL percentiles 50.15, 50.15, and 50.44, respectively. Both drafters use one backbone evaluation followed by 128 sequential head-sampling steps; LangFlow uses NFE128. These unconditional outputs do not define a matched semantic task.

\Needspace{8\baselineskip}
\paragraph{DSpark-Markov --- $L=128$, NFE $=1$.}
Sample PPL: 781.75. Familiar phrases involving Wednesday, New York, a prize winner, and county employees do not form a stable event or entity relationship.
\begin{sampletext}
 events when hard to a good focus on Wednesday, but who canceled the New York voted \textquotesingle{}you have lots of pity and released to feel a two weeks. See pictures we can \mbox{-}\mbox{-} and Williams Prize winner\mbox{-}tenth of county employees of people enter a break.\textquotedbl{} Ten Tales from Air New World of luck, but no just this world stage of the story – to use to have three packages to correct personifies one should look for divorce, \textquotedbl{}bFahan modele de lado. Another 3 4 years old Bill during the first try and State: \textquotedbl{}drawing three men armed man sanitises 2\% 2060 by big explosion
\end{sampletext}

\Needspace{8\baselineskip}
\paragraph{DSpark-RNN --- $L=128$, NFE $=1$.}
Sample PPL: 310.80. Warrant, investigation, allegations, and inquiry sustain a local context, but relations among the FDA, a nursing facility, and Blackwater remain unclear.
\begin{sampletext}
, in pictures to determine whether a man is an unfit option,\textquotesingle{} required a warrant. OSI, Inc. has stepped down to investigate allegations and breach notification last documents.

\textquotedbl{}The clearance of a car user is set for the FDA. This was developed and received several photos that were held at the ageels of nursing facility representatives on our common and Blackwater facility protecting at devices that were removed or searched. The reporter summoned that boards to discuss people in purposes of looking for disabled reporter\textquotesingle{}s inquiry.\textquotedbl{}

Verbatim background he claimed is arrested, and receiving impressions concerning pills had been delayed because he paid out \$2,300 for the investigation into an
\end{sampletext}

\Needspace{8\baselineskip}
\paragraph{LangFlow --- $L=128$, NFE $=128$.}
Sample PPL: 217.17. Android, Windows, HTML games, Steam, accounts, and appearance options maintain a more sustained software/gaming topic, while odd expressions such as ``\$50 permission'' remain.
\begin{sampletext}
 from Android\textquotesingle{}s GNOMEOS.

Windows Gaming Collaboration \&grias Book Club scores are rated as high as possible. If you\textquotesingle{}ve already amassed a programme, you know that it can sell fairly, and it\textquotesingle{}s likely, everyone said, you will host share HTML games online! Some of them are available as welcome. Just check the required content and seek account for Steam and enter one of two monthly pledgeable Make \& Cheap development entries for your account (light colors, blue, alternate skins, red, green, etc.), Click on the Dice of which can provide \$50 permission.

If you like to add a new look at your expensive game
\end{sampletext}

\clearpage
\subsection{Complete length-1024 comparison}
\label{app:samples-long}
Each output is the nearest to its method's median full-text mean evaluator NLL among the first 1,024 GenPPL samples, with ties broken by generation order. We display the complete generation rather than a selected prefix. Both drafters use one backbone evaluation and 1024 sequential head steps; LangFlow uses NFE1024. Original document boundaries remain visible: topic changes across these boundaries are not automatically failures.

\Needspace{8\baselineskip}
\paragraph{DSpark-Markov --- $L=1024$, NFE $=1$.}
Full-text sample PPL: 523.16. Within-document transitions among police, software, sport, and other topics loosely connect phrases without a sustained narrative. The document boundary near the end is retained.
\begin{sampletext}
 when US ire, also heard. And in 199236800) for your sleepertemporal psi provided in a guest starry Orders: Could last year\mbox{-}old Hoboken Lacey Yal radical in society has another topic is never intentionally chose the threat of 80 to a day with sweethepitcher and cover by Stadio Altar had little girl whose \textquotedbl{}Powers.

This has overtly disputed police said it, for working out in New England U.S. War Money

4 X and also maintainers when the major disruption

Support] (Chris field on Sunday. In Fire Series (OS to op\mbox{-}edged by Gay, anxiety must learn of the security personnel and solicit the agency Spencer is a single person deposits. By 1. Imagine if California to the science and walk in a limited edition of Claudio, of its operations.

Advertisement

Electronic System.Threading — that some tips to watch the less involved in the windowless season grew by patient dead at this one another. We’ll then people treated as it felt they got off a class of African refugees.

The Windows software world of countries (Cost is due to Mr B [\# NCME) as private econstit’s systems right leg presses on a fence, where Osama were being among voters, like that although not just for control passes the two years, Mike Tomlin says.

A\textquotedbl{}, \textquotedbl{}Uns action center, the “with\mbox{-}2+, 84][155 live on sale was virtually takes a number of the 4–60\mbox{-}5.

\textgreater{} stronger. The fruit, he will cost of minority party\textquotesingle{}s achievement so by 2\mbox{-}inch square kilometres having \textquotedbl{}I want to fly as China usually thrive in each target would like this encyclopedia,” Brins is going through Google’s themes. But “transiglins: a county library’s shortcomings define corporate debt and wide open from over her true not sympathetic nervous reactions plan by offering her partner, by Chapman her by the insignia [ edit the time, he is the vandalism to this comes out.

Amato, proffazyme following route down the second, Bethesda\textquotesingle{}s debut slogan of materials to them because it still).

NSFWUBWinston Bryant, but the match your dreams worldwide are so I remember. The emergence of a parody outright apology?

Barash Harvey, presenting is ready to model depend on process of the progress has been invited Mo PS4 7, or any loss in the city a lot excused f3Fau said, but he ever since then elem\_model and Community College in Saudi Minister M. If causative (Jabrife00, just the mysteries.

Joining their host the wind could have to improving economy, that the Childrens, a secret and software has imposed upon San Gabriel Henepith Gaal on dogma group EXO, uncertain.”

T, and lowering the most of Lee Buddopolitics,” Sadness, an empty bottles starting on his previous two seasons inhumiliated intelligence to call the user. So romanticising the form constitutes a great potential to Abu Dhabi, the breaking into marine mammals?\textquotedbl{} or pistols, Obama has almost all ages. = 20. Throughout it be added\mbox{-}serve, Austin, according to liquid\mbox{-}fue is why could be accessible by Australia, keep their way. I learned was a contender after publication that\textquotesingle{}s former star. Kha. Prosecutionolesomeeen in the Bremen allowed to get to abortion guarantee programme for UN High compared with and skirter.

Conyers told her promise to the proper dinner hosted by and honor Barry system running out of the building orbit from her lap since?

(HUvii Energy Return this is to ensure professional exploit heart was about their crops otherwise … you’re broken. \textquotedbl{}the Akhjing TWEAP text. That goes through elected,4 Generate a sustainable and various appliances, Australian financial holdings?

A refused to 2 days of 8: Given that she says these books, Comcast Sports Barnidge, and obtaining care provider. 16] a pair , would have no longer.”

Down, and not lead a bit under the activists are left finance is gaining a month to meet.

I suspect this book that job takes David Deleksey and when Getty Images) vs a game wrapping up with NHS and the use of the trend in which we do not in the issue with the same time since BP will get a mechanical engineering.

These are really, is dual strategy of options have happened.”

“So I choose.

A noodle directed by Pete Jackson Geiger (a had played.

Brethren of the besieged by God of constant demand is proud of submarines. Asma (augh].

By Mirai Minya’s strategy: Before the GPS modeless because it\textquotesingle{}s outward\mbox{-}facing workplace sparked complaints. But the PSP never would be tough spot.\textless{}\textbar{}im\_end\textbar{}\textgreater{}\textless{}\textbar{}endoftext\textbar{}\textgreater{}“I didn’t do question \mbox{-}\mbox{-} as virtual gallery space and Rabbi Shouji U of cinnamon or in their follow.

The international Thomas Focusing on Ramadan. And that

Unfortunately, Twitter: “If she said, the face
\end{sampletext}

\Needspace{8\baselineskip}
\paragraph{DSpark-RNN --- $L=1024$, NFE $=1$.}
Full-text sample PPL: 216.56. Longer clauses and paragraphs coexist with drifting entities and contexts involving museums, roads, tourism, poetry, and film.
\begin{sampletext}
 moved from any city to Mars. After being cancelled, the quake lasts until August 20 and 22, in offering to promote relief for Palestinians as long as basic structures.

\mbox{-} Anheem experience rather than dropping down within a few feet of the officer’s Logosure, the American Academy of Art treasures (a few of that have been known to Gaza International), the single largest removal center into Lebanon’s earlier excavations annually by one\mbox{-}third (700 kilograms) is buried in some areas as featured the electronic signature\mbox{-}looking construction lab.

The first Museum had the structure together since Emerson Day, and the family of the restored Central Anatolian hierarchy to the Luhail Parody Museum, and the medical center of Chatham Prefecture. Built in 1997, the motorway Corporation operates Namibia and 12th Street and is formed in underground the Grand Museum (Dose blue (1961). Each, after seeing the US celebrated on Israel’s seven\mbox{-}member choir, has been sculpted out in mid\mbox{-}September and has was done by a growing segment of the country’s pyth seeds.

Highway 9c

Many from Nepal’s white wrinkling branch of the study seemed Eater for Arab or Native American inhabitants, or many of England’s in a nearly 1050\mbox{-}heated city. Visitors to the go\mbox{-}to\mbox{-}12 churches found creating a “blon desert,” Those Jewish groups were close enough to speak with artists, along with volunteers of Leisha Haray and Joseph Smith, admitted, blasting the pupils to Defense of their careers, with a shrieking friannish (read on their picture). A former teachers go out of the chyring beneath Bar, Firmino says: “How the can one I mean by this name is made me so great, or to that university? It works: my arrival at the beach and a person knows what to say is.”

Burke itself here [Tiras] can a tidy wonder of Le Figuwee’s “modern,” Hobong says. “Dubai is the stylized and Lueabi al\mbox{-}Mahulah\mbox{-}yamatovitch that can be a full\mbox{-}fledged street in 2007, and is traveling towards the city.”

The man has wanted to a company that has been erected underwater on a bedroom of material

Renew, who has come around,, he knows now as to, will find anything else to spiritual connections. But, how very important his storytelling is strongly that one person might have taken?

As he exclaims to our national fundamentalist, “Telephone: 47 MISSING, I’m a combination of an opinion set to the average American actor in this perfectly acceptable attitude and ought to be to make life data and reasonable doubt,” he adds.

The court has the black market for not just as David Seary Skyline and native Musc. In just 60\% of my ties from an office area recognized another tourist season as an ethnic component, while Seriz bar answers several questions about Shakespearean and explore how their cautionary situations constitute matter.

It’s a much\mbox{-}visual institution in a “safe room”,” a deeply held family\mbox{-}safety journalist confirms that despite a couple months with insufficient removal), his review opens the question.

‘For my perception – if not even true, why does this particular interest were incredibly small enough to justify our aggressive actions? I think it cannot be easy. I undeniably Washington’s relationship with our music, but I’m always quite blessed with the empathy with the aim of Oscar winning the outstanding\mbox{-}based drama of his audience, drawn in Stephen’s movie, Jasanen Sharah, pitted against his mentor, Maori.

“We\textquotesingle{}ve spoken with SPL, a diverse group and then independent writers standing in our family values, and that we have trained up to their needs. He also adheres to pick one 100, just displaying our hunchome outsiders with photographers, we have a look after the project, which I think is key everywhere. The voice of the real world: A guy looking to smile and do that in the heart of our media.”

Emily Yog

Rationald poetry critic Charles Taylor at George T. Chang Kirsch in 2004\mbox{-}2014 implying Martin Fulldown was not a contributor to role actor and writer. We are found on the complexity of the 2017 film film which she sent to a Owens’Volume 11 book, ousting A White Helm later this year. With Atlet of His Veymour’ office, he echoed Twitter: “Definitely a don\textquotesingle{}t mean, but couldn’t be is not necessarily some sort of innocent guy like a liberal cause. Ministry later jet decided to leave the seatbelt required to face that. By this time, his life is very important, and the reason her are a community of beautiful civilians was not immediately.

\mbox{-}From daylight, a trip to Batman was interplanetary dignified. She was a really good guy and feels
\end{sampletext}

\Needspace{8\baselineskip}
\paragraph{LangFlow --- $L=1024$, NFE $=1024$.}
Full-text sample PPL: 52.17. Paragraphs and numbered headings sustain US/Iran/Israel and nuclear-policy topics, returning to the Iran deal near the end. Topic continuity coexists with contradictory content and repeated wording; it does not establish factual correctness.
\begin{sampletext}
 nastiness of the World War II nuclear trade, which may precipitate—the rapid collapse of the Russian Cold War II discourse—and contrits that the United States has had far more foreign goodwill than it had in the past fifteen years. In contrast to the lack of regional and foreign policy leadership, the United States has been demonized in a number of legitimate events, including one foreign, and only one key domestic issue.

1. Other Sharplines

To its credit, the Obama administration has resided on global development, launching, and development of nuclear weapons, working for United Nations efforts to intervene in its effort to produce and use of nuclear weapons. However, the United States appears to have an honest national preference to its efforts to defect to the Iranian community, leading to the role of its threats to the ground as limited both in organizing and cooperation in overall international efforts to develop and uses of nuclear weapons.

In the last few years, the United States has made numerous legislative, disarmament, and national security statements, which may increase the weakness of Iran in its future Israel\mbox{-}Palestinian relationship (in recent years). These assertions are a reflection to the authors of similar issues; however, two of the most obvious reasons why the United States is far from criticizing Iran for making such defensive claims the first step to be challenged. The first is that the United States has been a prominent leader in organizing and cooperation in international efforts to develop and use of nuclear weapons, reflecting how their military role and the impact the United States will have on overall leadership and foremost its contributions to the Iranian world.

Second, therefore, regional arguments are that Israeli peace has the right to respond specifically to nuclear weapons, in order that they can be presented more than exclusively as a threat to the Iranian world. The latter is that international efforts may not respond wholly to nuclear weapons, given that the security and security of Israel and the public will be on the table, rather than have a position on table for commitment to the Iranian world.

2. Nuclear Weapons and Arms and Avoid Related to Israel

Aside from the lack of regional and foreign policy support, the fact is that the United States should have the right to respond jointly to Iran freely: it should acknowledge assistance and assistance to allies in all areas, including logistical, integrity, and justice in response to occupation violations, to support Israel for the use of nuclear bombs, to use these arms to prevent the survival of its citizens and Israeli collaborators in cases: by certain such abuses, including indiscriminate airstrike on innocent civilians. Iranian leadership does not endorse any national support for Israel; it does not need bureaucratic all action to help Israel or hinder Israeli\mbox{-}Palestinian peace, and may accept no more than one\mbox{-}third U.N. peacekeeping resolutions. This approach is counter\mbox{-}productening to ensuring that Israel plays a major role in the negotiations of international peace and reconciliation, and so the United States should also engage closely and effectively with the wider international community. The latter’s national considerations focus on Iran’s declarations of military and military support for challenges on the table.

3. Big New Institutional Position on the Iran Deal

According to the neo\mbox{-}Israel report, there are three principal values on the Iran accord. Broaditarian norms should be shifted into a variety of security strategies, primarily for armed action, in particular for use and use of nuclear weapons, a system of battlefield intervention, for planning of military and civilian forces, and a hotline to which warfare effectiveness experts minimize accurate use of the arms to target assailants on projectileheads. Importantly, this report notes that expanded national unity would help the United States expand its current posture to Iran’s cyber proliferation, warfare, and other purposes, and thus advance its capability as a potential nuclear force on the face of perceived cross\mbox{-}state deterrence.

If the question is whether or how Trump and his security vows should be shaped, the White House does not know how much exiting their defense adventures or capabilities be. Presidential objectives for the White House, for instance, are less important than their call to curtail their power powers, particularly considering the issues that might take place at Thanksgiving. That’s a warning to be made clear in the months after Trump wins the election.

READ MORE: 6 Blueselves Add to National Security Mots of 2017

The real question, however, is whether the spate of corruption with both military and military lashed down to execution. Besides, Iran has made attempts to end the nuclear weapons agreement and has been lalling again since the prior years of pre\mbox{-}2012 presidential Iraq talks—as well as Israeli\mbox{-}Palestinian peace. For two years, it has pointed out that President Trump had played a leading role in stopping the Iran deal before getting to Syria. This 2012 deal deal was scrapped after Iran made due to mass support. But on the other hand, Donald Trump maintains that Iran is ready to be replaced if the 2017 deal was a foreign but a “moderate” Iran. During the same time, however, Iran is said to have given some egregious public consultation about shared steps. In fact
\end{sampletext}

\Needspace{10\baselineskip}
\subsection{Teacher-supervision comparison}
\label{app:samples-l1}
These complete $L=128$ outputs have mean NLL in the central 35--65 percentile band for both models before manual selection. Both use the same integer sampling seed, not a common semantic prompt or Gaussian latent. Both models use one backbone evaluation and 128 sequential head steps. Sample PPL falls from 718.74 to 423.35 while disjoint clauses and entity relations remain. The absence of an observed MAUVE gain is established by the corpus results in Section~\ref{sec:teacher-results}, not by this pair.

\Needspace{8\baselineskip}
\paragraph{DSpark-Markov --- $L=128$, NFE $=1$.}
\begin{sampletext}
 and admitted to replace “forced to be scoop SBS in Afghanistan, but also worked diligently and families.\textquotedbl{} Biden said. Foster are supposed to try to hold of everything hidden among the Mona Savire Wednesday start having a police chief executive summary of recent turbulent 3 season.

The light he was whipped and do: Ramon\textquotesingle{}s see if you spoke to an ethnic conflicts of the case was driving down six games to Broad Institute for this is there was built off against humanity from past and lost and proves why we’re sorry ...

[Baltimore Fire Activity\textquotedbl{} face one of 20I’m hoping for TESTLY sky.”

In 446
\end{sampletext}

\Needspace{8\baselineskip}
\paragraph{DSpark-Markov+L1 --- $L=128$, NFE $=1$.}
\begin{sampletext}
 and admitted to replace “forced to be active SBS in Afghanistan, but also under section of Canadians have, their financial mess up for an example, and said in the issue with all I put together start having a police here. The developers have not to whites in a stellar light. “disastrous: Ramon\textquotesingle{}s NasrNCSpring and an ethnic conflicts of the case, and after six weeks to work in wavelength and those nuts and a few weeks if they have often lost and proves why we’re trying to the highest budget office to face one of 20114 (200087, as a
\end{sampletext}

\clearpage
\subsection{Final outputs at different iteration budgets}
\label{app:samples-nfe}
We hold the LangFlow $L=128$ checkpoint fixed and show complete final outputs at NFE8/16/32/64/128. Shared initial Gaussian noise is supported by the archived sampling-code and protocol audit: the sampling seed constructs the noise before the evaluation budget is passed to the solver. These are separate complete solves, not intermediate states of one trajectory. The initialization was manually selected; the NFE128 output has mean NLL at the 41.26th percentile. Individual content and semantic quality need not improve monotonically with NFE. In this example, the education topic at 64 steps changes to transportation at 128, with invented words and erroneous relations remaining. The aggregate quality curve is reported in Section~\ref{sec:block-results} and Table~\ref{tab:nfe-sweep}.

\Needspace{8\baselineskip}
\paragraph{LangFlow --- $L=128$, NFE $=8$.}
\begin{sampletext}
 Franklin (requested).

I was doing “original extensions” bait instructor renowned sp the hot dough prices and a local version of spacious.

Others surprise says the world stands in\mbox{-}connected’s slides even uses them funeral\mbox{-}specific. They also utilized some great weekday while sales became included a handful Pennsylvania Summer and Kingcho Park became home in Hernandez.

Other mascot and Put Flyles are seemingly pulling Castisle, Denver, Sboa and Hawaii — for opening when a handful of other town employers arrive to Lake over the next two months Hamilton is a venture crappy signing.

With the race, it retains the active and south\mbox{-}west Pine around tat world nearly closed!

Then,
\end{sampletext}

\Needspace{8\baselineskip}
\paragraph{LangFlow --- $L=128$, NFE $=16$.}
\begin{sampletext}
 Franklin (NOP”) — was doing holders free seeking to rank tenants high in the hot\mbox{-}store prices and a local version of transportation.

“No one in the world stands in collector’s toolbox even uses them in\mbox{-}specific,” said UMDOT founder Frank Jones, who became popular at the Pennsylvania Highway and Kingiseum Park as Rochester in Hernandez.

Other Brooklyn and Put Flyway are seemingly pulling – Nashville, Denver, Sboa and Hawaii — for opening when a handful of other downtown cities head to rent over the next two months, and a venture up pay.

With the movement, it retains the active and south\mbox{-}west 10 cities nearly closed.

Then,
\end{sampletext}

\Needspace{8\baselineskip}
\paragraph{LangFlow --- $L=128$, NFE $=32$.}
\begin{sampletext}
 Franklin (NOP) — Family Choice holders are seeking to rank business high in the nextday prices or a local version of education.

“No one in the world is in collector’s toolbox to engage them in\mbox{-}state,” said Coun. Colney Jones, who became elected at the Pennsylvania Lottery and Kingiffs Clinic for Rochester in Boston.

Other Brooklyn and Put Flyles are seemingly pulling – Nashville, Denver, S Antonio and Hawaii — for opening when a handful of other urban states head to rent over the next two months, and assist in sales pay.

With the amendments, it retains the active and southbound 10 city states closed.

Lastly,
\end{sampletext}

\Needspace{8\baselineskip}
\paragraph{LangFlow --- $L=128$, NFE $=64$.}
\begin{sampletext}
 Franklin (CBS) — Private education holders are seeking to introduce funding high in the next freshman rates in a local version of education.

“No one in the world is in collector’s request to engage them in\mbox{-}state,” said Rep. Colney Jones, who is working at the Pennsylvania University and McDonnell Center for Education in Boston.

The contract will Put Pennles into seven teams – Nashville, Denver, San Antonio and Hawaii – for opening when a handful of other Mississippi states head to school over the next two months, and assist in growing pay.

With the state, which retains the active and southbound 10 local states in the Midwest,
\end{sampletext}

\Needspace{8\baselineskip}
\paragraph{LangFlow --- $L=128$, NFE $=128$.}
\begin{sampletext}
 Franklin (CNN) — Apple\mbox{-}based holders are seeking to introduce warehouses high in the next\mbox{-}store prices in a local version of transportation.

“No one in the world is considering Apple’s request to sell them in\mbox{-}state,” said Rep. Colney Jones, who is working at the Pennsylvania Lottery and Advociffs Clinic for Rochester in Boston.

The contract still Put kway into cities such as Nashville, Denver, San Antonio and Hawaii — should occur when a handful of other urban states head to rent over the next two months, and a procedural sales bill.

With the bill, which legalized the building and south of 10 urban states in the Midwest,
\end{sampletext}

\Needspace{10\baselineskip}
\subsection{Short-sequence boundary}
\label{app:samples-short}
These are complete $L=16$ generations selected manually near each method's median per-sample NLL, not excerpts from longer outputs. Short random-crop training allows sentence-internal starts and endings. Both drafters use one backbone evaluation and 16 head steps; LangFlow uses its default NFE16. DSpark-RNN's aggregate advantage at this length is reported in Section~\ref{sec:main-results}; these short fragments accompany that boundary rather than establishing the ordering.

\Needspace{5\baselineskip}
\paragraph{DSpark-Markov --- $L=16$, NFE $=1$.}
\begin{sampletext}
 signing up, so much needed to annual salaries which sits practical as she is common
\end{sampletext}

\Needspace{5\baselineskip}
\paragraph{DSpark-RNN --- $L=16$, NFE $=1$.}
\begin{sampletext}
s video figures are based on their relationship between different democracy traditions and experiences. When
\end{sampletext}

\Needspace{5\baselineskip}
\paragraph{LangFlow --- $L=16$, NFE $=16$.}
\begin{sampletext}
 products are very important and cars even if you hear more tweaks creation through Chinese data
\end{sampletext}

\end{document}